\documentclass{article} 

\usepackage{booktabs}
\usepackage{algorithm}
\usepackage{algpseudocode}
\usepackage{caption}
\usepackage{graphicx}
\usepackage{tikz}
\usepackage{multirow}
\usepackage{adjustbox}
\usepackage{amsmath}
\usepackage{amsfonts}
\usepackage{bm}

\newcommand\Appref[1]{Appendix~\ref{#1}}
\newcommand\Tabref[1]{Table~\ref{#1}}

\newcommand{\ts}{{\mX}}

\newcommand{\diff}{{\mathrm{d}}}
\newcommand{\noise}{\bm{\epsilon}}
\newcommand{\diffusionContInd}{{s}}
\newcommand{\diffusionContLast}{{S}}
\newcommand{\diffusionInd}{{n}}
\newcommand{\diffusionLast}{{N}}
\newcommand{\timeSeriesSize}{{M}}
\newcommand{\scoreNet}{{\psi_\vtheta}}
\newcommand{\score}{{\nabla_{\vx_\diffusionContInd}\! \log p(\vx_\diffusionContInd)}}
\newcommand{\Ver}{\parallel}

\def\Figref#1{Figure~\ref{#1}}

\def\Secref#1{Section~\ref{#1}}

\def\eqref#1{equation~\ref{#1}}
\def\Eqref#1{Equation~\ref{#1}}

\def\1{\bm{1}}

\def\vzero{{\bm{0}}}

\def\vmu{{\bm{\mu}}}
\def\vtheta{{\bm{\theta}}}
\def\va{{\bm{a}}}

\def\vm{{\bm{m}}}

\def\vq{{\bm{q}}}

\def\vt{{\bm{t}}}

\def\vx{{\bm{x}}}

\def\vz{{\bm{z}}}

\def\mA{{\bm{A}}}

\def\mI{{\bm{I}}}

\def\mK{{\bm{K}}}
\def\mL{{\bm{L}}}

\def\mQ{{\bm{Q}}}
\def\mR{{\bm{R}}}

\def\mX{{\bm{X}}}

\def\mZ{{\bm{Z}}}

\def\mSigma{{\bm{\Sigma}}}

\DeclareMathAlphabet{\mathsfit}{\encodingdefault}{\sfdefault}{m}{sl}
\SetMathAlphabet{\mathsfit}{bold}{\encodingdefault}{\sfdefault}{bx}{n}

\newcommand{\E}{\mathbb{E}}

\newcommand{\R}{\mathbb{R}}

\newcommand{\KL}{D_{\mathrm{KL}}}

\usepackage[backref=page]{hyperref}
\hypersetup{
  colorlinks,
  linkcolor={red!50!black},
  citecolor={blue!50!black},
  urlcolor={blue!80!black}
}

\usepackage[accepted]{icml2023}

\icmltitlerunning{Modeling Temporal Data as Continuous Functions with Stochastic Process Diffusion}

\begin{document}

\twocolumn[
    \icmltitle{Modeling Temporal Data as Continuous Functions\\with Stochastic Process Diffusion}

    \begin{icmlauthorlist}
        \icmlauthor{Marin Bilo\v{s}}{tum,ms}
        \icmlauthor{Kashif Rasul}{ms}
        \icmlauthor{Anderson Schneider}{ms}
        \icmlauthor{Yuriy Nevmyvaka}{ms}
        \icmlauthor{Stephan G\"unnemann}{tum}
    \end{icmlauthorlist}
    
    \icmlaffiliation{tum}{Technical University of Munich, Germany}
    \icmlaffiliation{ms}{Machine Learning Research, Morgan Stanley, United States}
    
    \icmlcorrespondingauthor{Marin Bilo\v{s}}{marin.bilos@tum.de}
    
    \icmlkeywords{Machine Learning}
    
    \vskip 0.3in
]

\printAffiliationsAndNotice{}

\begin{abstract}
Temporal data such as time series can be viewed as discretized measurements of the underlying function. To build a generative model for such data we have to model the stochastic process that governs it. We propose a solution by defining the denoising diffusion model in the function space which also allows us to naturally handle irregularly-sampled observations. The forward process gradually adds noise to functions, preserving their continuity, while the learned reverse process removes the noise and returns functions as new samples. To this end, we define suitable noise sources and introduce novel denoising and score-matching models. We show how our method can be used for multivariate probabilistic forecasting and imputation, and how our model can be interpreted as a neural process.
\end{abstract}

\vspace{-0.5cm}
\section{Introduction}\label{sec:introduction}

Time series data is collected from measurements of some real-world system that evolves via some complex unknown dynamics. The sampling rate is often arbitrary and non-uniform, producing irregularly-sampled time series.
Therefore, we can make an assumption that time series follows some underlying continuous function; consider, e.g., the temperature or load of a system over time. Although values are observed as separate events, we know that temperature always exists and its evolution over time is smooth, not jittery. This kind of data can be found in many domains, from medical, industrial to financial applications.

Previously, different approaches for modeling irregular data have been proposed, including neural ordinary and stochastic differential equations \citep{chen2018neural,li2020scalable}, neural processes \citep{garnelo2018neural}, normalizing flows \citep{deng2020modeling}, etc. As it turns out, capturing the true generative process proves difficult, especially with the inherent stochasticity of the data.

Recently, denoising diffusion models have shown great promise in modeling very complicated data distributions such as those in the image domain \citep{ho2020denoising,song2021sde}. The approach consists of first gradually adding the noise to data, until it becomes pure noise, corresponding to some base distribution. At the same time, the model is trained to reverse this process. To generate a new data point, we start with an initial noisy value sampled from the base distribution; then the model gradually denoises it to reach a sample from the learned data distribution. We define this more rigorously in \Secref{sec:background}.

In this work, we expand on this general framework to define the diffusion for data measured in continuous time by treating it as a discretization of some continuous function. 
That is, instead of adding noise to each data point independently, we add the noise to the whole function while preserving its continuity. In \Secref{sec:diffusion}, we show that this can be done by using stochastic processes as noise generators. We additionally show that the final noisy function will also correspond to a sample from a known stochastic process. Next, we specify the transition probabilities in the forward noising process, the evidence bound on the likelihood used in the training, and the new sampling procedure, for both the fixed-step and SDE-based diffusion approaches.

\begin{figure*}[!t]
    \centering
    \vspace{-0.2cm}
    \includegraphics[width=\textwidth]{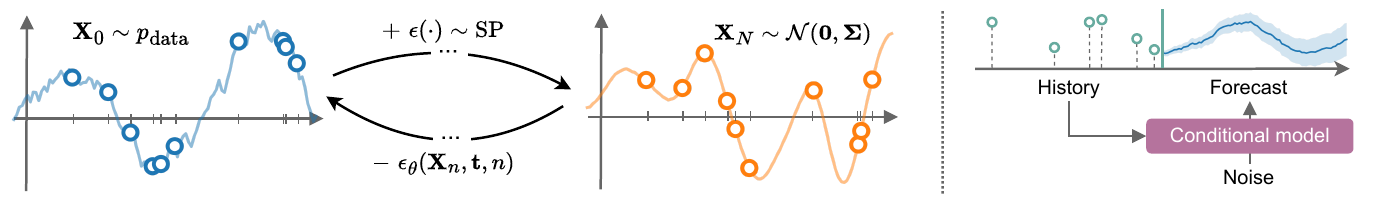}
    \vspace{-0.7cm}
    \caption{(Left) We add noise from a stochastic process (SP) to the \emph{whole} time series at once. The model $\noise_\theta$ learns to reverse this process. (Right) We can use this approach to, e.g., forecast with uncertainty.}
    \label{fig:fig1}
\end{figure*}

\Figref{fig:fig1} shows an illustration of our approach. Data is observed as a set of (irregularly-sampled) points that correspond to some underlying function. By adding noise to this function we reach the prior stochastic process. At the same time, the model can reverse this process, allowing us to generate new function samples.

In \Secref{sec:model} we describe different use cases that we tackle with our model while highlighting the benefits over previous approaches. For instance, we use conditioning, to output the distribution over future values, i.e., for multivariate probabilistic forecasting. Since we define the distribution over functions we can also view our model as a neural process \citep{garnelo2018neural}, allowing us to estimate missing points from the observed. In \Secref{sec:experiments} we empirically show that our model outperforms the baselines on all tasks.

\section{Background}\label{sec:background}

Generally, given training data $\{\vx_i\}$, with $\vx_i \in \R^d$, the goal of generative modeling is to learn the probability density function $p(\vx)$ and be able to generate new samples from this learned distribution. Diffusion models achieve both of these goals by learning to reverse some fixed process that adds noise to the data.
In the following, we present a brief overview of the two ways to define  diffusion; in \Secref{sec:discrete} the noise is added across $\diffusionLast$ increasing scales \cite{ho2020denoising}, which is then taken to the limit in \Secref{sec:continuous_diffusion} using a stochastic differential equation (SDE) \cite{song2021sde}.

\subsection{Fixed-step diffusion}\label{sec:discrete}

\citet{pmlr-v37-sohl-dickstein15, ho2020denoising} propose the denoising diffusion probabilistic model (DDPM) which gradually adds \emph{fixed} Gaussian noise to the observed data point $\vx_0$ via known scales $\beta_\diffusionInd$ to define a sequence of progressively noisier values $\vx_1, \vx_2, \dots, \vx_\diffusionLast$. The final noisy output $\vx_\diffusionLast \sim \mathcal{N}(\vzero, \mI)$ carries no information about the original data point. The sequence of positive noise (variance) scales $\beta_1, \dots, \beta_\diffusionLast$ has to be increasing such that the first noisy output $\vx_1$ is close to the original data $\vx_0$, and the final value $\vx_\diffusionLast$ is pure noise. The goal is then to learn to reverse this process.

As diffusion forms a Markov chain, the transition between any two consecutive points is defined with a conditional probability $q(\vx_\diffusionInd | \vx_{\diffusionInd-1}) = \mathcal{N}(\sqrt{1-\beta_\diffusionInd} \vx_{\diffusionInd-1}, \beta_\diffusionInd \mI)$.  Since the transition kernel is Gaussian, the value at any step $\diffusionInd$ can be sampled directly from $\vx_0$. Let $\alpha_\diffusionInd = 1 - \beta_\diffusionInd$ and $\bar{\alpha}_\diffusionInd = \prod_{k=1}^\diffusionInd \alpha_k$, then we can write:
\begin{align}\label{eq:q_xi_x0}
    q(\vx_\diffusionInd | \vx_0) = \mathcal{N}(\sqrt{\bar{\alpha}_\diffusionInd} \vx_0, (1 - \bar{\alpha}_\diffusionInd) \mI).
\end{align}
Further, the probability of any intermediate value $\vx_{\diffusionInd-1}$ given its successor $\vx_\diffusionInd$ and initial $\vx_0$ is
\begin{align}\label{eq:q_posterior}
    q(\vx_{\diffusionInd-1} | \vx_\diffusionInd, \vx_0) = \mathcal{N}(\tilde{\vmu}_\diffusionInd, \tilde\beta_\diffusionInd \mI),
\end{align}
\begin{flalign*}
    \text{where:} && &\tilde{\vmu}_\diffusionInd = \frac{\sqrt{\bar{\alpha}_{\diffusionInd-1}} \beta_\diffusionInd}{1 - \bar{\alpha}_\diffusionInd} \vx_0 + \frac{\sqrt{\alpha_\diffusionInd} (1 - \bar\alpha_{\diffusionInd-1})}{1 - \bar\alpha_\diffusionInd} \vx_\diffusionInd, &\\
    && &\tilde\beta_\diffusionInd = \frac{1 - \bar{\alpha}_{\diffusionInd-1}}{1 - \bar{\alpha}_\diffusionInd} \beta_\diffusionInd . &
\end{flalign*}
The generative model learns the reverse process. To this end, \citet{pmlr-v37-sohl-dickstein15} set $p(\vx_{\diffusionInd-1} | \vx_\diffusionInd) = \mathcal{N}(\vmu_\theta(\vx_\diffusionInd, \diffusionInd), \beta_\diffusionInd \mI)$, and parameterized $\vmu_\theta$ with a neural network. The training objective is to maximize the evidence lower bound, $\log p(\vx_0) \geq$
\begin{align}\label{eq:elbo_ddpm}
\begin{split}
    \E_q[& 
        \log p(\vx_0 | \vx_1)
        - \KL(q(\vx_\diffusionLast | \vx_0) \| p(\vx_\diffusionLast)) \\
        - &\textstyle \sum_{\diffusionInd > 1} \KL(q(\vx_{\diffusionInd-1} | \vx_\diffusionInd, \vx_0) \| p(\vx_{\diffusionInd - 1} | \vx_\diffusionInd))
    ].
\end{split}
\end{align}
In practice, however, the  approach  by \citet{ho2020denoising} is to reparameterize $\vmu_\theta$ and predict the noise $\noise$ that was added to $\vx_0$, using a neural network $\noise_\theta(\vx_\diffusionInd, \diffusionInd)$, and minimize the simplified loss function:
\begin{align}\label{eq:loss_ddpm}
    \mathcal{L} = \E_{\noise,\diffusionInd}
    \left[\lVert 
    \noise_\theta(\sqrt{\bar\alpha_\diffusionInd} \vx_0 + \sqrt{1 - \bar\alpha_\diffusionInd} \noise, \diffusionInd) 
    - \noise \rVert_2^2\right],
\end{align}
where the expectation is over $\noise \sim \mathcal{N}(\vzero, \mI)$ and $\diffusionInd \sim \mathcal{U}(\{ 0, \dots, \diffusionLast \})$.
To generate new data, the first step is to sample a point from the final distribution $\vx_\diffusionLast \sim \mathcal{N}(\vzero, \mI)$ and then iteratively denoise it using the above model ($\vx_\diffusionLast \mapsto \vx_{\diffusionLast -1} \mapsto \dots \mapsto \vx_0$) to get a sample from the data distribution.
To summarize, the forward process adds the noise $\noise$ to the input $\vx_0$, at different scales, to produce $\vx_\diffusionInd$. The learned model inverts this, i.e., predicts $\noise$ from $\vx_\diffusionInd$.

\subsection{Score-based SDE diffusion}\label{sec:continuous_diffusion}

Instead of taking a finite number of diffusion steps as in \Secref{sec:discrete}, \citet{song2021sde} introduce a continuous diffusion of vector valued data, $\vx_0 \mapsto \vx_\diffusionContInd$ where $\diffusionContInd \in [0, \diffusionContLast]$ is now a continuous variable. The forward process can be elegantly defined with an SDE:
\begin{align}\label{eq:diffusion_sde}
    \diff \vx_\diffusionContInd = f(\vx_\diffusionContInd, \diffusionContInd) \diff \diffusionContInd + g(\diffusionContInd) \diff W_\diffusionContInd,
\end{align}
where $W$ is a standard Wiener process. The variable $\diffusionContInd$ is the continuous analogue of the discrete steps implying that  the input gets noisier during the SDE evolution. The final value $\vx_\diffusionContLast \sim p(\vx_\diffusionContLast)$ will follow some predefined distribution, as in \Secref{sec:discrete}.
For the forward SDE in \Eqref{eq:diffusion_sde} there exist a corresponding reverse SDE \citep{anderson1982reverse}:
\begin{align*}
    \diff \vx_\diffusionContInd = [f(\vx_\diffusionContInd, \diffusionContInd) - g(\diffusionContInd)^2 \score] \diff \diffusionContInd + g(\diffusionContInd) \diff W_\diffusionContInd,
\end{align*}
where $\score$ is the score function. Solving the above SDE from $\diffusionContLast$ to $0$, given  initial condition $\vx_\diffusionContLast \sim p(\vx_\diffusionContLast)$, returns a sample from the data distribution. The generative model's goal is to learn the score function via a neural network $\scoreNet(\vx_\diffusionContInd, \diffusionContInd)$, by minimizing:
\begin{align}\label{eq:continuous_diffusion_loss}
    \mathcal{L} = 
    \E_{\vx_\diffusionContInd,\diffusionContInd}
    \left[\lVert 
        \scoreNet(\vx_\diffusionContInd, \diffusionContInd) - \score
    \rVert_2^2 \right] ,
\end{align}
with $\vx_\diffusionContInd \sim \mathrm{SDE}(\vx_0)$ and $\diffusionContInd \sim \mathcal{U}(0, \diffusionContLast)$.
\citet{song2021sde} define the continuous equivalent to DDPM forward process as the following SDE:
\begin{align}\label{eq:ddpm_continuous}
    \diff \vx_\diffusionContInd = -\frac{1}{2} \beta(\diffusionContInd) \vx_\diffusionContInd \diff \diffusionContInd + \sqrt{\beta(\diffusionContInd)} \diff W_\diffusionContInd,
\end{align}
where $\beta(\diffusionContInd)$ and $\diffusionContLast$ are chosen in such a way that ensures the final noise distribution is unit normal, $\vx_\diffusionContLast \sim \mathcal{N}(\vzero, \mI)$. Given this specific parameterization, one can easily derive the transition probability $q(\vx_\diffusionContInd | \vx_0)$ and calculate the exact score in closed-form (see \Secref{sec:ts_continuous_diffusion} and \Appref{app:continuous_diffusion_transition}).

\subsection{Extensions}

Generative modeling with diffusion recently gained traction as it provides good sampling quality on image generation \citep{dhariwal21diffusion,ramesh22dalle,stablediffusion} and became the state-of-the-art method replacing GANs \citep{goodfellow2020generative}. The modeling power translates to other tasks as well, so it has been used in, e.g., modeling waveforms \citep{KongPHZC21} and time series forecasting \citep{rasul2021autoregressive}, but also generating discrete data such as text \citep{AustinJHTB21} and molecules \citep{anand2022protein,lee2022proteinsgm}. In this work we tackle a different task---generating continuous functions.
Many of the advances over the original diffusion focused on improving the sampling speed \citep{chung2022come,jolicoeur2021gotta,lyu2022accelerating}, while others implement the noise scheduling for better modeling capacity \citep{
nichol2021improved,kingma2021variational}. This area of research is orthogonal to our proposed method as we can easily implement any of the techniques that improve general diffusion, to make our method perform faster or have better sampling quality.

\section{Diffusion for time series data}\label{sec:diffusion}

In contrast to the previous section which deals with data points that are represented by vectors, we are interested in  generative modeling for  time series data.
We represent the data as a time-indexed sequence of points observed across $\timeSeriesSize$ timestamps: $\ts = ( \vx(t_0), \dots, \vx(t_{\timeSeriesSize-1}) )$, $t_i \in \vt \subset [0, T]$. The observations can be equally spaced but this formulation encompasses  irregularly-sampled data as well. We assume that each observed time series comes from its corresponding underlying continuous function $\vx(\cdot)$.

Our approach can be viewed as modeling the distribution ``$p(\vx(\cdot))$" over functions instead of  vectors, which amounts to learning the stochastic process. We review stochastic processes in more detail in \Secref{sec:diffusion_neural_process}.
To preserve continuity, we cannot apply the ideas from \Secref{sec:background} directly, unless we assume measurements are independent of each other. One issue of adding independent noise in the diffusion arises because it produces discontinuous samples.

\subsection{Stochastic processes as noise sources for diffusion}\label{sec:ts_diffusion_noise}

Instead of defining the diffusion by adding some scaled  noise vector $\noise \sim \mathcal{N}(\vzero, \mI)$ to a data vector $\vx$, we add a noise \emph{function} (stochastic process) $\noise(\cdot)$ to the underlying data function $\vx(\cdot)$. The only restriction on $\noise(\cdot)$ is that it has to be continuous so that the output remains continuous as well, which clearly rules out stochastic processes where time is indexed by a \emph{finite} set, e.g., $\noise( \vt) \sim \mathcal{N}(\vzero, \mI)$. However, using a normal distribution proved to be very convenient in \Secref{sec:background} as it allowed for closed-form formulations of various terms, especially the loss. This is due to the nice properties that Gaussian random variables have.

\looseness=-1
Therefore, our goal is to define $\noise(\cdot)$ which will satisfy the continuity property while giving us tractable training and sampling. Note that $t$ refers to the time of the observation and $\noise(t)$ is the noise at $t$, in contrast to the previous section where \emph{time-like} variables $\diffusionInd$ and $\diffusionContInd$ referred to the noise scale.

We could consider obtaining the noise from a standard Wiener process $\noise(t) = W_t$. A clear disadvantage of this approach is that variance grows with time. Additionally, the distribution of $W_0$ is degenerate as  we never add any noise. This can be solved in an ad hoc manner by shifting the whole time series similar to \citet{deng2020modeling}.

\looseness=-1
Instead, in the following, we present two \emph{stationary} stochastic processes that add the same  amount of noise regardless of the time of the observation. Note that the noise is  \emph{correlated} in the time dimension, hence the use of the stochastic process. An additional nice property of these processes is that they reduce to the diffusion from \Secref{sec:background} in the trivial case of time series with only one element.

Let us shortly restrict the discussion to univariate time series $\ts \in \R^{\timeSeriesSize}$ and producing noise $\noise(\vt) \in \R^{\timeSeriesSize}$. We present the general approach at the end of this section.

\textbf{A)~Gaussian process prior.} Given a set of $\timeSeriesSize$ time points $\vt$, we propose sampling $\noise(\vt)$ from a Gaussian process $\mathcal{N}(\vzero, \mSigma)$ where each element of the covariance matrix is specified with a kernel $\mSigma_{ij} = k(t_i, t_j)$, where $t_i, t_j \in \vt$. This produces \emph{smooth} noise functions $\noise(\cdot)$ that can be evaluated at any $t$.
To define a stationary process, we have to use a stationary kernel; we will use a radial basis function $k(t_i,t_j) = \exp(-\gamma (t_i - t_j)^2)$. Adjusting the parameter $\gamma$ (or $\sigma = 1 / \gamma$) lets us vary the flatness  of the noise curves. 
Given a set of time points $\vt$, we can easily sample from this process by first computing the covariance $\mSigma(\vt)$ and then sample from the multivariate normal distribution $\mathcal{N}(\vzero, \mSigma)$.

\textbf{B)~Ornstein-Uhlenbeck diffusion.} An alternative noise distribution is a stationary OU process that is specified as a solution to the following SDE:
\begin{align}\label{eq:ou_sde}
    \diff\epsilon_t = -\gamma \epsilon_t \diff t + \diff W_t,
\end{align}
where $W_t$ is the standard Wiener process and we use the initial condition $\epsilon_0 \sim \mathcal{N}(0, 1)$. We can obtain samples from OU process easily by sampling from a time-changed and scaled Wiener process: $\smash{ \exp(-\gamma t) W_{\exp(2 \gamma  t)} }$. The covariance can be calculated as $\smash{\mSigma_{ij} = \exp(-\gamma |t_i - t_j|)}$.
The OU process is a special case of a Gaussian process with a Mat\'ern kernel ($\nu = 0.5$) \citep[p.\ 86]{williams2006gaussian}. We discuss different sampling techniques for OU process and their trade-offs in \Appref{app:ou_sampling}.

In the end, both the GP and OU processes are defined with a multivariate normal distribution over a finite collection of points, where the covariance is calculated using the times of the observations. As opposed to the methods from \Secref{sec:background}, we use correlated noise in the forward process. Our approach allows us to produce continuous functions as samples and will prove to be a natural way to do forecasting and imputation.

\textbf{Multivariate time series.}
In our work, we consider multivariate time series which means we observe an evolution of a $d$-dimensional vector over time. In the forward diffusion process, we treat the data as $d$ individual univariate time series and add the noise to them independently. This is equivalent to using block-diagonal covariance matrix of size $(\timeSeriesSize d)\times(\timeSeriesSize d)$ with $\mSigma$ repeated on the diagonal. This is in line with the previous works where, e.g., independent noise is added to individual pixels in an image.

Note that this does not mean we do not model correlations between dimensions. As we will see in the following section, the reverse process takes a complete multivariate time series and captures these correlations. This is again similar to image synthesis---although forward process is independent over pixels, the reverse process captures the whole image. The difference in our approach is that we also enforce the continuity across the time dimension, which means our model is guaranteed to produce continuous samples.

\subsection{Discrete stochastic process diffusion (DSPD)}\label{sec:ts_discrete_diffusion}

We apply the discrete diffusion framework to the time series setting. Note, \emph{discrete} refers to the number of diffusion steps (\Secref{sec:discrete}), i.e., we still model continuous functions. Reusing the notation from before, $\ts_0$ denotes the input data and $\ts_\diffusionInd = ( \vx_\diffusionInd(t_0), \dots, \vx_\diffusionInd(t_{\timeSeriesSize-1}) )$ is the noisy output after $\diffusionInd$ diffusion steps. 
In contrast to the classical DDPM \cite{ho2020denoising} where one adds independent Gaussian noise to data, we now add the noise from a stochastic process. In particular, given the times of the observations, we can compute the covariance $\mSigma$ and sample noise $\noise(\cdot)$ from a GP or OU process as defined in \Secref{sec:ts_diffusion_noise}. We write the transition kernel and the posterior as:
\begin{align}
    q(\ts_\diffusionInd | \ts_0) &= \mathcal{N}(\sqrt{\bar{\alpha}_\diffusionInd} \ts_0, (1 - \bar{\alpha}_\diffusionInd) \mSigma), \label{eq:ts_q_xn_x0}\\
    q(\ts_{\diffusionInd-1} | \ts_\diffusionInd, \ts_0) &= \mathcal{N}(\tilde{\vmu}_\diffusionInd, \tilde\beta_\diffusionInd \mSigma) . \label{eq:ts_q_posterior}
\end{align}
We provide a full derivation  in \Appref{app:ddpm_posterior}. Even though we are now able to sample functions instead of single points, the distributions are still similar to the previous case with the only change occurring in the covariance. This nice result will be useful later to define the loss which is analogous to \Eqref{eq:loss_ddpm}.

We define the generative model as the reverse process $p(\ts_{\diffusionInd - 1} | \ts_\diffusionInd) = \mathcal{N}(\vmu_\theta(\ts_\diffusionInd, \vt, \diffusionInd), \beta_\diffusionInd \mSigma)$, similar to \citet{ho2020denoising}, keeping the time-dependent covariance $\mSigma$. The key difference is that the model now takes the full time series consisting of noisy observations $\ts_\diffusionInd$ with their timestamps $\vt$ in order to predict the noise $\noise$ which has the same size as $\ts_\diffusionInd$. The architecture, therefore, has to be a type of a time series encoder-decoder.

Since each distribution that appears in the ELBO (\Eqref{eq:elbo_ddpm}) is multivariate normal, the loss can be calculated in closed-form. In \Appref{app:ddpm_loss} we show the full derivation. Further, we show how we can reparameterize the model such that the covariance $\mSigma$ disappears from the final loss. In particular, if our model predicts the noise that was added to the original data we can simplify the loss to only compute the squared difference between the predicted and true noise, similar to \Eqref{eq:loss_ddpm}:
\begin{align}\label{eq:ddpm_continuous_loss}
\mathcal{L} =
    \E_{\noise,\diffusionInd}
    \left[\lVert 
    \noise_\theta(\sqrt{\bar\alpha_\diffusionInd} \ts_0 + \sqrt{1 - \bar\alpha_\diffusionInd} \noise, \vt, \diffusionInd) - \noise \rVert_2^2\right].
\end{align}
Finally, in order to sample, the initial noise has to come from a stochastic process instead of an independent normal distribution. The same is the case for the noise that is used in the intermediate steps of the Langevin dynamics. We show the implementation of training (Algorithm~\ref{alg:training_ddpm}) and sampling (Algorithm~\ref{alg:sampling_ddpm}) for Gaussian process diffusion in \Appref{app:algorithms}. The Ornstein-Uhlenbeck case is analogous---we simply change the noise source.

\subsection{Continuous stochastic process diffusion (CSPD)}\label{sec:ts_continuous_diffusion}

Similarly to the previous section, we can extend the continuous diffusion framework to use the noise coming from a Gaussian or OU process. Now, the noise scales $\beta(\diffusionContInd)$ are continuous in the diffusion time $\diffusionContInd$, see \Secref{sec:continuous_diffusion}. Given a factorized covariance matrix $\mSigma = \mL \mL^T$, we modify the variance preserving diffusion SDE \citep{song2021sde}:
\begin{align}\label{eq:ts_continuous_diffusion_sde}
    \diff \ts_\diffusionContInd = -\frac{1}{2} \beta(\diffusionContInd) \ts_\diffusionContInd \diff\diffusionContInd + \sqrt{\beta(\diffusionContInd)} \mL \diff W_\diffusionContInd ,
\end{align}
which gives us the following transition probability:
\begin{align}\label{eq:ts_continuous_diffusion_transition}
    q(\ts_\diffusionContInd | \ts_0) = \mathcal{N}(\tilde\vmu, \tilde\mSigma) ,
\end{align}
with:
\begin{align}
\begin{split}
    \tilde\vmu &= \ts_0 e^{-\frac{1}{2} \int_0^\diffusionContInd \beta(s) \diff s}\\
    \tilde\mSigma &= \mSigma \left(1 - e^{-\int_0^\diffusionContInd \beta(s) \diff s}\right) .
\end{split}
\end{align}
This result is derived using Equation~5.51 from \citet{sarkka2019applied}, similar to an analogous result in \citet{song2021sde}. We discuss this in more detail in \Appref{app:continuous_diffusion_transition}.
Since this probability is normal, the value of the score function can be computed in closed-form:
\begin{align}\label{eq:ts_continuous_diffusion_score}
    \nabla_{\ts_\diffusionContInd} \log q(\ts_\diffusionContInd | \ts_0) = -\tilde\mSigma^{-1}(\ts_\diffusionContInd - \tilde\vmu) ,
\end{align}
which we can use to optimize the same objective as in \Eqref{eq:continuous_diffusion_loss}. Our neural network $\noise_\theta(\ts_\diffusionContInd, \vt, \diffusionContInd)$ will take in the full time series, together with the observation times $\vt$ and the diffusion time $\diffusionContInd$, and predict the values of the score function. As it turns out, we can again use the reparameterization in which we predict the noise, whilst the score is only calculated when sampling new realizations. That is, we represent the score as $\mL \tilde{\noise} / \sigma^2$, where $\sigma^2 = 1 - \exp(-\int_0^s\beta(s) \diff s)$ (\Eqref{eq:ts_continuous_diffusion_score}) and $\tilde\noise$ is the noise coming from an independent normal distribution.

\subsection{Related work}

Recently, \citet{kerrigan2022diffusion} proposed a similar approach for modeling functions with diffusion by defining a Gaussian measure on Hilbert spaces. This formalizes some of our ideas using the results from measure theory. Another related work \cite{dutordoir2022neural} views diffusion on functions as neural processes, similar to our formulation in \Secref{sec:diffusion_neural_process}. A concurrent work \cite{phillips2022spectral} uses KL decomposition to approximate the data in spectral space and then learns a standard diffusion model in this space. On the other hand, our method is well suited for irregular time series data, e.g., it naturally offers conditioning on observed data and performing imputation.

\section{Applications}\label{sec:model}

To train a generative model, it must learn to reverse the forward diffusion process by predicting the noise that was added to the clean data. The input to the model is the time series $(\ts_0, \vt)$ along with the diffusion step $\diffusionInd$ or diffusion time $\diffusionContInd$, and the output is of the same size as $\ts_0$.
If additional inputs are available, we can also model the conditional distribution; e.g., we often have covariates for each  time point of $\vt$. We can also condition the generation on the past observations which essentially defines a probabilistic forecaster (\Secref{sec:forecasting}) or condition only on the observed values which defines a neural process (\Secref{sec:diffusion_neural_process}) or an imputation model (\Secref{sec:imputation}).

\subsection{Forecasting multivariate time series}\label{sec:forecasting}

Forecasting is answering what is going to happen, given what we have seen, and as such is the most prominent task in time series analysis. Probabilistic forecasting adds the layer of (aleatoric) uncertainty on top of that and returns the confidence intervals which is often a requirement for deploying  models in  real world settings. The neural forecasters are usually encoder-decoders, where the history of observations $(\ts^H, \vt^H)$ is represented with a single vector $\vz$ and the decoder outputs the distribution of the future values $\ts^F$ given $\vz$ at time points $\vt^F$ . Previous works relied on specifying the parameters of the output distribution, e.g., via a diagonal covariance \citep{SALINAS20201181}, its low-rank approximation \citep{NIPS2019_8907}, normalizing flows \citep{NEURIPS2020_1f47cef5, rasul2021multivariate}, or GANs \citep{engproc2021005040}.

Recently, \citet{rasul2021autoregressive} introduced a diffusion-based forecasting model to learn the conditional probability $p(\ts^F | \ts^H)$, where $\ts^H = ( \vx(t_0), \dots, \vx(t_{\timeSeriesSize-1}) )$ is a history window of size $\timeSeriesSize$ sampled randomly from the full training data. They specify the distribution $p(\vx(t_{\timeSeriesSize}) | \ts^H)$ using a conditional DDPM model. The forward process adds independent Gaussian noise to $\vx(t_{\timeSeriesSize})$ the same way as in DDPM. However, the reverse denoising model is conditioned on the history $\ts^H$ which is represented with a fixed sized vector $\vz$. After training is completed, the predictions are made in the following way: (1) $\ts^H$ is encoded with an RNN to obtain $\vz$; (2) the initial noisy value is sampled $\vx_\diffusionLast(t_{\timeSeriesSize}) \sim \mathcal{N}(\vzero, \mI)$; and (3) denoising is performed using the sampling algorithm from \citet{ho2020denoising} but conditioned on $\vz$ to obtain $\vx_0(t_{\timeSeriesSize})$. The final denoised value is the forecast and sampling multiple values allows computing empirical confidence intervals of interest.

In \citet{rasul2021autoregressive}, the timestamps are always discrete and the prediction is autoregressive, i.e., the values are produced \emph{one by one}. Our diffusion framework offers these key improvements: (a) the predictions can be made at \emph{any} future time point, i.e., in continuous time, not discrete steps; and (b) we can predict \emph{multiple} values in parallel which scales better on modern hardware.

In our case, the prediction $\ts^F$ will not be a single vector but a sequence $( \vx(t_{\timeSeriesSize}), \dots, \vx(t_{\timeSeriesSize+K}) )$ of size $K$, where $K$ can vary in size. This type of data is naturally handled by our stochastic process diffusion as defined in \Secref{sec:diffusion}. Note that the predicted values are also not conditionally independent but we model the interactions between them in the denoising model $\noise_\theta$.

\begin{figure}[t]
    \centering
    \includegraphics[width=\linewidth]{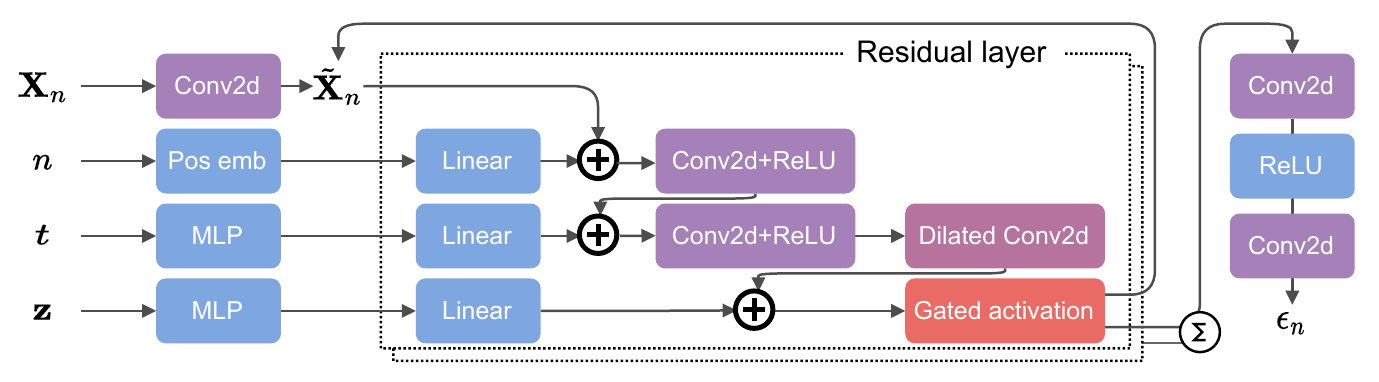}
    \caption{Overview of the forecasting model. The inputs are the noisy time series $\ts_\diffusionInd$, diffusion steps $\diffusionInd$, observation times $\vt$, and the history vector $\vz$. The output is the predicted noise value $\noise_\diffusionInd$. We use two-dimensional convolutions where \emph{height} and \emph{width} correspond to feature and time dimensions.}
    \label{fig:forecast_model}
    \vspace{-0.5cm}
\end{figure}

We design $\noise_\theta$ in the following way. Previous observations are represented with an RNN to obtain $\vz$ and condition the reverse process. We propose an architecture similar to the TimeGrad model \cite{rasul2021autoregressive,KongPHZC21}, but not autoregressive as it outputs all the values simultaneously. \Figref{fig:forecast_model} shows an overview. The model inputs noisy future values $\ts^F_\diffusionInd$, diffusion step $\diffusionInd$, future timestamps $\vt$ and the encoded history $\vz$. In contrast to previous works, we use 2D convolution where the extra dimension corresponds to the time dimension.

After training a DSPD-GP (\Secref{sec:ts_discrete_diffusion}), we can forecast:
\begin{enumerate}
    \item Encode the history $\ts^H$ with an RNN to get $\vz$,
    \item Sample initial prediction $\ts^F_\diffusionLast$ from a GP prior,
    \item Denoise using $\noise_\theta(\ts^F_\diffusionInd, \vt, \diffusionInd, \vz)$ with Algorithm~\ref{alg:sampling_ddpm}.
\end{enumerate}
Instead of an RNN we can also use transformers \cite{vaswani2017attention} but we wanted to keep the architecture similar to \citet{rasul2021autoregressive} and showcase the novel stochastic process-based diffusion.

\subsection{Diffusion process as a neural process}\label{sec:diffusion_neural_process}

So far we have used stochastic processes as noise sources to generate continuous functions. We can view such a model as a stochastic process as well. Stochastic process is defined as a collection of random variables $\{ X(t) \}_t$ indexed over some set $\mathcal{T}$, in our case $\mathcal{T} \subseteq \R$. We usually care about the finite sequences of points since this is what we encounter in our data. In that case, the model that defines some probability measure $p$ is a stochastic process if it satisfies consistency conditions, as defined in the Kolmogorov extension theorem \cite{oksendal2013stochastic}. Crucially, the model has to be permutation equivariant, i.e., the order of the incoming points should not matter.

Based on this, neural processes \citep{garnelo2018neural} are a class of latent variable models that define a stochastic process with neural networks. Given a set of data points (a dataset), the model outputs the probability distribution over the functions that generated this dataset. That is, for different datasets, the model will define different stochastic processes. Due to this behavior, neural processes bear a resemblance to the Gaussian processes but can also be viewed as a meta learning model \citep{hospedales2021meta}. 

Let $\ts^{A}$ denote the observed data, in our case, a time series, and let $\ts^{B}$ be the unobserved data at the time points $\vt^B$. \citet{garnelo2018neural} construct the encoder-decoder model that uses an amortized variational inference for training \citep{kingma2013auto}. The encoder takes in a set of observed points $(\ts^{A}, \vt^A)$ and outputs the distribution over the latent variable $q(\vz)$. The decoder takes in the sampled latent vector $\vz$ and the query time points $\vt^B$ and predicts the values of the unobserved points $\ts^B$. In order to produce permutation equivariant measure, it is crucial that the encoder is permutation invariant, i.e., the input order does not alter the result. Then the probability of $\ts^{B}$ is conditionally independent given $\vz$ \cite{de1937prevision}. This is easy to achieve using, e.g., deep sets \citep{zaheer2017deep}. 

Since our approach samples functions, we can condition the generation on an input dataset $(\ts^{A}, \vt^A)$ in order to create our version of a neural process, based purely on the diffusion framework. The encoder will be a deterministic neural network that outputs the latent vector $\vz$, contrary to \citet{garnelo2018neural} which outputs the distribution.  Similar to \Secref{sec:forecasting}, the diffusion is conditioned on $\vz$ and we can output samples for any query $\vt^B$. For example, if we again take DSPD-GP model, we sample as follows:
\begin{enumerate}
    \item Permutation invariant encode $(\ts^{A}, \vt^A)$ to get $\vz$,
    \item Sample initial points $\ts^{B}_\diffusionLast$ at $\vt^B$ from a GP prior,
    \item Denoise using $\noise_\theta(\ts^B_\diffusionInd, \vt^B, \diffusionInd, \vz)$ with Algorithm~\ref{alg:sampling_ddpm}.
\end{enumerate}

Therefore, we capture the distribution $p(\ts^B | \ts^A)$ directly. 

We achieve equivariance using a transformer-like model $\noise_\theta$ \cite{vaswani2017attention} that uses a learnable RBF kernel for a similarity function. The architecture is described in more detail in \Appref{app:neural_process}.
During training, we adopt the approach of feeding in data such that we learn $p(\ts^{A} \cup \ts^B | \ts^A)$ which helps our model output high certainty around $\vt^A$, see \citet{garnelo2018neural}.

In the end, our model sees many observed-unobserved pairs corresponding to different true underlying processes. The model learns to represent the observed points $\ts^A$ such that the denoising process corresponds to the correct distribution, given $\ts^A$. After training is completed, we take a time series $\ts^A$ and output samples at any set of query time points $\vt^B$. We can view such an approach as an interpolation or imputation model that fills-in the missing values across time. The main appeal is the ability to capture different stochastic processes within a single model.

A similar idea by \citet{dutordoir2022neural} proposes using diffusion as an alternative to Gaussian processes, however, it uses an independent noise, therefore, it does not guarantee producing continuous functions.

\subsection{Probabilistic time series imputation}\label{sec:imputation}

The previous section considered interpolating  in time. Now, we look into  filling-in the missing values across the observation dimensions, i.e., the imputation of the vectors. Each element $\vx(t_i)$ of the time series $\ts$ is assigned a mask $\vm$ of the same dimension that indicates whether the $j$-th value $x^{(j)}$ of the vector $\vx(t_i)$ has been observed ($m^{(j)}=1$) or if it is missing ($m^{(j)}=0$).

Given observed $\ts^A$ and missing points $\ts^B$, \citet{tashiro2021csdi} propose a model that learns a conditional distribution $p(\ts^B | \ts^A)$. The model is built upon a diffusion framework and the reverse process is conditioned on $\ts^A$, similar to that in \Secref{sec:diffusion_neural_process}.
We extend this by introducing noise from a stochastic process, as presented above. The learnable model remains the same but we introduce the correlated noise in the loss and sampling. We posit that continuous noise process, as an inductive bias for the irregular time series, is a more natural choice.

\section{Experiments}\label{sec:experiments}

\subsection{Probabilistic modeling}

We start by investigating pure generative capabilities of our model, i.e., unconditional generation of time series.\footnote{\scriptsize \url{https://github.com/morganstanley/MSML/tree/main/papers/Stochastic_Process_Diffusion}}

\textbf{Baselines.} Previously, neural ODEs \citep{chen2018neural} were introduced as a way to capture the irregularly sampled time series since they can naturally handle the continuous time. As such, they can be seen as a building block that can also be used alongside our method to devise different denoising networks.
\citet{rubanova2019latent} construct an encoder-decoder architecture based on neural ODEs which resembles the variational autoencoder \citep{kingma2013auto}. The time series is, thus, modeled in a latent space by sampling a random vector which is propagated with an ODE. Neural SDEs \citep{li2020scalable} extend this by adding noise in every solver step but they either do not produce noisy-enough samples \citep{li2020scalable} or use an adversarial objective which is difficult to train \citep{kidger2021neural}.
Finally, continuous-time flow process  (CTFP) \citep{deng2020modeling} uses normalizing flows \citep{kobyzev2020normalizing} to generate the time series by sampling the initial noise from the stochastic process and transform it with an invertible function to obtain the sample from the target distribution. Although this allows exact likelihood training, the method cannot capture some processes \citep{deng2021continuous} and is often augmented to be trained as a VAE.

\textbf{Data.} We generate 6 synthetic datasets, each with 10000 samples, that involve stochastic processes, dynamical and chaotic systems. CIR (Cox-Ingersoll-Ross) is the stochastic differential equation often used in finance,  Lorenz is a chaotic system in three dimensions, OU is generated using a specific parameterization of \Eqref{eq:ou_sde}, Predator-prey and Sink are two-dimensional dynamical systems, and Sine is generated as a mixture of random sine waves. Full details on generation are included in \Appref{app:probabilistic_modeling}.

\textbf{Ablation.}
We test our DSPD and CSPD with independent Gaussian noise and noise from a stochastic process (GP and OU) on the above described datasets.
We first check whether using a model that captures interactions across time (e.g., RNN or transformer) outperforms the model that treats each data point in the time series independently. \Tabref{tab:further_discriminator_results} (\Appref{app:probabilistic_modeling}) shows we need to model the interaction across time, as expected.

\begin{figure}[!t]
    \centering
    \resizebox{0.95\linewidth}{!}{
    \input{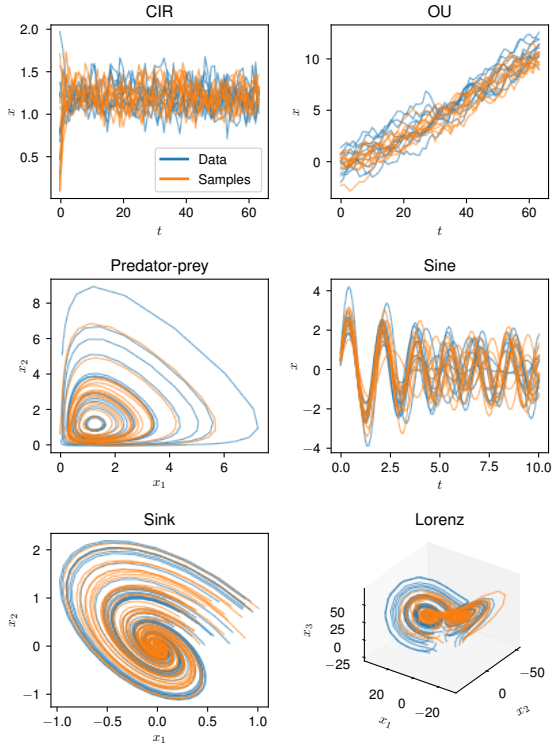}
    }
    \vspace{-0.3cm}
    \caption{Real data (in blue) and samples from our model (in orange) based on diffusion with Gaussian process noise.}
    \label{fig:synthetic_samples}
    \vspace{-0.15in}
\end{figure}

Now, we check if having a stochastic process noise is better than the independent Gaussian noise, i.e., we compare our method to \citet{ho2020denoising,song2021sde}. 
\Tabref{tab:synthetic_results} (\Appref{app:probabilistic_modeling}) shows that using a stochastic process achieves lower negative log-likelihood. We report the results only for CSPD model as it allows likelihood evaluation, whereas DSPD returns ELBO. The results for ELBO are similar. The gap between stochastic and independent noise is especially evident on datasets where we need to generate complicated samples.  The difference is less visible in \emph{noisy} datasets, such as CIR, but our method shows much better performance when generating smooth curves such as Lorenz. Finally, \Figref{fig:synthetic_samples} demonstrates the quality of the samples.

\begin{table}[!t]
    \centering
    \caption{Accuracy of the discriminator trained to distinguish real data and model samples (closer to 0.5 is better).}
    \label{tab:discriminator_results}
    \vspace{0.15cm}
    \begin{adjustbox}{width=\linewidth}
        \begin{tabular}{lccc}
            & CTFP & Latent ODE & DSPD-GP (Our) \\
\midrule
CIR         & 0.998$\pm$0.001 & 1.0$\pm$0.0 & \textbf{0.511$\pm$0.028} \\
Lorenz      & 0.995$\pm$0.006 & 0.998$\pm$0.002 & \textbf{0.513$\pm$0.028} \\
OU          & 0.783$\pm$0.076 & \textbf{0.512$\pm$0.033} & \textbf{0.505$\pm$0.045} \\
Predator-prey &0.789$\pm$0.023 & 0.958$\pm$0.021 & \textbf{0.585$\pm$0.022} \\
Sine        & 0.981$\pm$0.01 & 1.0$\pm$0.0 & \textbf{0.525$\pm$0.009} \\
Sink        & 0.726$\pm$0.138 & 0.907$\pm$0.039 & \textbf{0.513$\pm$0.01} \\
\end{tabular}
    \end{adjustbox}
    \vspace{-0.5cm}
\end{table}

\textbf{Results.}
Since likelihood cannot be evaluated for all models and is not the best indication of the final sampling quality, we report the discriminative score.
That is, we quantitatively compare the generative power of our model with the established baselines for irregular time series modeling, namely, latent ODEs \citep{rubanova2019latent} and CTFP \citep{deng2020modeling} (details in \Appref{app:probabilistic_modeling}) by comparing the quality of the samples. In short, after training a single generative model we sample new data from it. The original and generated data is then used to train a new model that learns to discriminate between them. We then report the performance of a discriminator on the held-out test set. If the discriminator cannot be trained, i.e., its prediction is not better than a random guess, we say the generative model captures the true distribution.

\Tabref{tab:discriminator_results} compares our model with the baselines and demonstrates that we produce samples that are indistinguishable to a powerful transformer-based discriminator. The same does not hold for the competing methods.

Note that we also implemented the latent SDE model \cite{li2020scalable} and we observe that latent ODE outperforms it, which is why we do not include it in the main results. 

We notice differences in sampling times for different methods. In particular, CTFP is the fastest followed by our method and neural ODEs, which have similar runtime.
Even though diffusion requires evaluating the denoising network $\diffusionLast$ times, ODE approaches oversample in the time dimension. Another difference is that neural ODEs slow down as they learn more complex dynamics since adaptive solver takes more steps. Same can be true for continuous version of diffusion models. The exact times depend on the architecture and hyperparameters as well as other design choices. For example, we present different ways to sample from OU process in \Appref{app:ou_sampling}. Depending on the choice we can trade-off having low memory impact (option 2) or ability to parallelize (option 3). We would like to highlight that, for all datasets in this paper, using stochastic process noise did not significantly impact performance compared to independent noise.

\subsection{Forecasting}

\begin{figure}[t]
    \centering
    \resizebox{\linewidth}{!}{
        \input{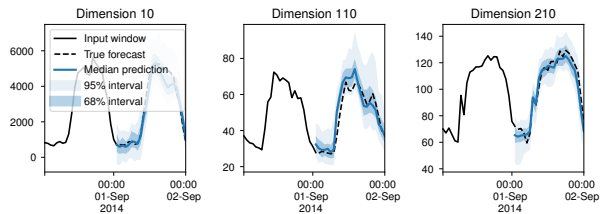}
    }
    \captionof{figure}{Forecast and uncertainty intervals on Electricity.}
    \label{fig:forecasting}
\end{figure}

We test our model as defined in \Secref{sec:forecasting} and \Figref{fig:forecast_model} against TimeGrad \citep{rasul2021multivariate} on three established real-world datasets: Electricity, Exchange and Solar \citep{lai2018modeling}. Due to the limitations of the CRPS-sum metric \citep{koochali2022random}, we report the NRMSE and the energy score \citep{doi:10.1198/016214506000001437} averaged over five runs, but we note that the rank of the model's performance does not change when using other metrics as well. For completeness, we include other time series baselines such as Gaussian process forecaster \cite{salinas2019high} in \Tabref{tab:forecasting_crps} (\Appref{app:forecasting}). \Tabref{tab:forecasting} shows that our method outperforms TimeGrad even though we predict over the complete forecast horizon at once, and \Figref{fig:forecasting} demonstrates the prediction quality alongside the uncertainty estimate.

\begin{table}[t]
    \centering
    \caption{NRMSE (top rows) and energy score (bottom rows) on real-world forecasting data, averaged over five runs.}
    \label{tab:forecasting}
    \vspace{0.1cm}
    \begin{tabular}{lcc}
& TimeGrad & Ours \\
\midrule
\multirow{2}{*}{Electricity} & 0.064$\pm$0.007 & \textbf{0.045$\pm$0.002} \\
& 8425$\pm$613 	&  \textbf{7079$\pm$164} \\ 
\midrule
\multirow{2}{*}{Exchange} & 0.013$\pm$0.003 & \textbf{0.012$\pm$0.001} \\
& 0.057$\pm$0.002 & \textbf{0.031$\pm$0.002} \\ 
\midrule
\multirow{2}{*}{Solar} & 0.799$\pm$0.096 & \textbf{0.757$\pm$0.026} \\
& \textbf{150$\pm$17} & 166$\pm$12 \\ 
\end{tabular}
\end{table}

\subsection{Neural process} 

\begin{figure}[t]
    \centering
    \resizebox{\linewidth}{!}{
        \input{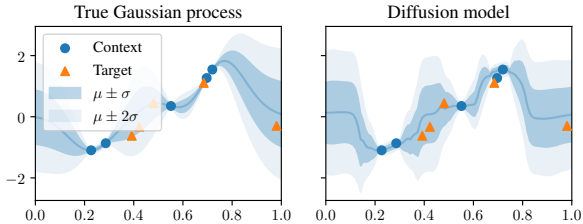}
    }
    \caption{Sampled curves given a set of points.}
    \label{fig:neural_process}
\end{figure}

We construct a dataset where each time series $\ts$ comes from a different stochastic process, by sampling from Gaussian processes with varying kernel parameters and time series lengths. This is a standard training setting in neural process literature \citep{garnelo2018neural}. In our denoising network, we modify the attention-like layer to make it stationary (see \Appref{app:neural_process}) and train as described in \Secref{sec:diffusion_neural_process}.
Due to the use of $\tanh$ activations in the final layers, combined with its stationary, our model extrapolates well, i.e., when $\tanh$ saturates the mean and the variance fall to zero-one. This is the same behaviour we see in the GP with an RBF kernel, for example.
The quantile loss of the unobserved data under the true GP model is $0.845$ while we achieve $0.737$ which indicates we capture the true process, which can also be seen in \Figref{fig:neural_process}. We remark that the attentive neural process \citep{kim2019attentive} does not produce the correct uncertainty. 

\begin{figure}[t]
    \centering
    \resizebox{\linewidth}{!}{
        \input{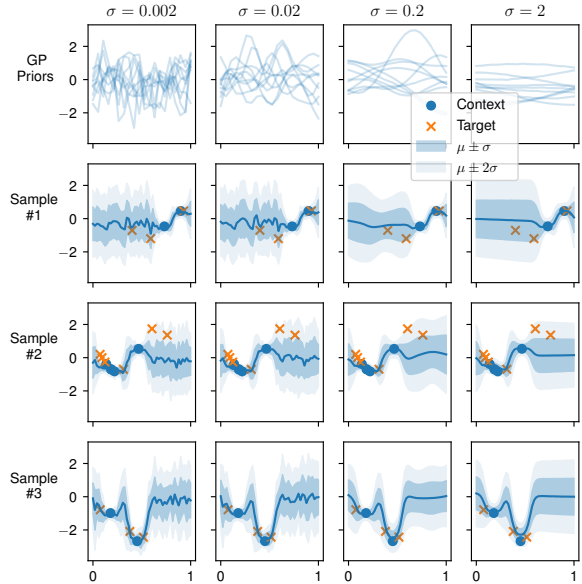}
    }
    \caption{Neural process with Gaussian process diffusion, fitted on GP synthetic data. Columns correspond to different values of the kernel parameter $\sigma = 1/\gamma$. The first row shows samples from the GP prior. As we can see, the higher the value of $\sigma$ the smoother the process will be. This is also reflected in the samples from the model. We show the same for OU process in \Figref{fig:ou_np}.}
    \label{fig:gp_np}
\end{figure}

Finally, in \Figref{fig:gp_np} we show how model behaves across different kernels. The noise process is connected to the final sample \emph{smoothness} but not the marginal distribution which are always correctly captured.

\subsection{Imputation} 

We compare to the CSDI \citep{tashiro2021csdi}, introduced in \Secref{sec:imputation}, on an imputation task. To this end, we use exactly the same training setup, including the random seeds and model architecture, but change the noise source to a Gaussian process. Following \citet{tashiro2021csdi}, we use Physionet dataset \citep{silva2012predicting} which is a collection of medical time series collected at an hourly rate. It already contains missing values but for testing purposes, we choose varying degrees of missingness and report the results on the test set. We update the loss and sampling accordingly, as in \Secref{sec:diffusion}. \Tabref{tab:csdi} shows that we outperform the original CSDI model even though we only changed the noise, and the dataset we used has regular time sampling. In \Appref{app:imputation} we provide more details, including the Wilcoxon one-sided signed-rank test \citep{conover1999practical} that shows our results have statistical significance.

\begin{table}[t]
    \caption{Imputation RMSE on Physionet with varying amounts of missingness. See \Appref{app:imputation} for more results.}
    \label{tab:csdi}
    \vspace{0.1cm}
    \centering
    \begin{tabular}{rcc}
Missing & CSDI & DSPD-GP (Our) \\
\midrule
10\%  & 0.520$\pm$0.055 & \textbf{0.498$\pm$0.036} \\
50\%  & \textbf{0.644$\pm$0.024} & \textbf{0.644$\pm$0.029} \\
90\%  & 0.818$\pm$0.02 & \textbf{0.815$\pm$0.019} \\
\end{tabular}

\end{table}

\section{Discussion}\label{sec:discussion}

In this paper, we introduced a novel generative model for continuous functions. It can also be viewed as a neural stochastic process or a generative model for solutions to stochastic differential equations. We also demonstrate how it can be used in conventional (both regular and irregularly-sampled) time series tasks such as forecasting, interpolation, and imputation.
In the experiments we showed that the improvements over the previous works come from using the stochastic process as the noise source; and using the model that takes in the whole time series at once.
The results demonstrate the practical utility of our method and validate our motivation.

\subsection{Future work}

We used bare bones diffusion without extensive tuning to demonstrate the modeling potential and make a fair comparison to other methods. However, it should be straightforward to improve upon our models by implementing recent advances in diffusion models \citep[e.g.,][]{nichol2021ddpmpp}. 
In case we have a large number of points, we can consider replacing the current sampling strategies with more scalable variants, such as switching to a sparse Gaussian processes \citep{candela05inducing}.
We can also explore different architecture choices, e.g., implement improvements in conditioning models via learned activations \citep{ramos2022conditioning}.
Finally, we can also apply the presented methods to other areas outside the time series domain, such as modeling point clouds or images, as we have demonstrated that our method is competitive on regular grids.

\section*{Acknowledgements}

M.B.\ completed part of the work while interning at Morgan Stanley. M.B.\ and S.G.\ are supported by the German Federal Ministry of Education and Research (BMBF), grant no. 01IS18036B.

\bibliography{references.bib}
\bibliographystyle{icml2023}

\newpage
\onecolumn
\appendix
\section{Derivations}

\subsection{Discrete diffusion posterior probability}\label{app:ddpm_posterior}

We extend \citet{ho2020denoising} by using full covariance $\mSigma(\vt)$ to define the noise distribution across time $\vt$. If $\mSigma = \mL \mL^T$ and keeping the same definitions from \Secref{sec:discrete} for $\beta_\diffusionInd$, $\alpha_\diffusionInd$, and $\bar\alpha_\diffusionInd$, we can write:
\begin{align}
    \ts_\diffusionInd &= \sqrt{1 - \beta_\diffusionInd} \ts_{\diffusionInd - 1} + \sqrt{\beta_\diffusionInd} \mL \noise, \\
    \ts_\diffusionInd &= \sqrt{\bar\alpha_\diffusionInd} \ts_0 + \sqrt{1 - \bar\alpha_\diffusionInd} \mL \noise,
\end{align}
with $\noise \in \mathcal{N}(\vzero, \mI)$. This corresponds to the following transition distributions:
\begin{align}
    q(\ts_\diffusionInd | \ts_{\diffusionInd - 1}) &= \mathcal{N}(
        \sqrt{1 - \beta_\diffusionInd} \ts_{\diffusionInd - 1},
        \beta_\diffusionInd \mSigma
    ),\\
    q(\ts_\diffusionInd | \ts_0) &= \mathcal{N}(
        \sqrt{\bar\alpha_\diffusionInd} \ts_0,
        (1 - \bar\alpha_\diffusionInd) \mSigma
    ).
\end{align}
We are interested in $q(\ts_{\diffusionInd - 1} | \ts_\diffusionInd, \ts_0) \propto q(\ts_\diffusionInd | \ts_{\diffusionInd - 1}) q(\ts_{\diffusionInd - 1} | \ts_0)$. Since both distributions on the right-hand side are normal, the result will be normal as well. 
We can write the resulting distribution as $\mathcal{N}(\tilde{\vmu}, \tilde{\mSigma})$, where:
\begin{align*}
    \tilde{\vmu} &= \mR (\ts_\diffusionInd - \mA \vmu_1) + \vmu_1 \\
    \tilde{\mSigma} &= \mSigma_1 - \mR \mA \mSigma_1^T \\
    \mR &= \mSigma_1 \mA^T (\mA \mSigma_1 \mA^T + \mSigma_2)^{-1} ,
\end{align*}
with 
$\mA = \sqrt{1 - \beta_\diffusionInd} \mI$, 
$\vmu_1 = \sqrt{\bar\alpha_{\diffusionInd - 1}} \ts_0$,
$\mSigma_1 = (1 - \bar\alpha_{\diffusionInd - 1}) \mSigma$, and
$\mSigma_2 = \beta_\diffusionInd \mSigma$.
We can now write:
\begin{align*}
    \mR &= (1 - \bar\alpha_{\diffusionInd - 1}) \mSigma \sqrt{1 - \beta_\diffusionInd} \left(\sqrt{1 - \beta_\diffusionInd} (1 - \bar\alpha_{\diffusionInd - 1}) \mSigma \sqrt{1 - \beta_\diffusionInd} + \beta_\diffusionInd \mSigma \right)^{-1} \\
    &= \frac{(1 - \bar\alpha_{\diffusionInd - 1})\sqrt{\alpha_\diffusionInd}}{\alpha_\diffusionInd (1 - \bar\alpha_{\diffusionInd - 1}) + 1 - \alpha_\diffusionInd} \mSigma \mSigma^{-1} \\
    &= \frac{1 - \bar\alpha_{\diffusionInd - 1}}{1 - \bar\alpha_\diffusionInd} \sqrt{\alpha_\diffusionInd},
\end{align*}
and from there:
\begin{align}\label{eq:q_posterior_mean_derivation}
\begin{split}
    \tilde{\vmu} &= \frac{1 - \bar\alpha_{\diffusionInd - 1}}{1 - \bar\alpha_\diffusionInd} \sqrt{\alpha_\diffusionInd}
    \left( 
        \ts_\diffusionInd - \sqrt{1 - \beta_\diffusionInd} \sqrt{\bar\alpha_{\diffusionInd - 1}} \ts_0
    \right)
    + \sqrt{\bar\alpha_{\diffusionInd - 1}} \ts_0 \\
    &= \frac{1 - \bar\alpha_{\diffusionInd - 1}}{1 - \bar\alpha_\diffusionInd} \sqrt{\alpha_\diffusionInd} \ts_\diffusionInd
    + 
    \sqrt{\bar\alpha_{\diffusionInd - 1}} 
    \left( 
        1 - \frac{1 - \bar\alpha_{\diffusionInd - 1}}{1 - \bar\alpha_\diffusionInd} \alpha_\diffusionInd
    \right)
    \ts_0 \\
    &= \frac{1 - \bar\alpha_{\diffusionInd - 1}}{1 - \bar\alpha_\diffusionInd} \sqrt{\alpha_\diffusionInd} \ts_\diffusionInd
    + 
    \frac{\sqrt{\bar\alpha_{\diffusionInd - 1}}}{1 - \bar\alpha_\diffusionInd} \beta_\diffusionInd \ts_0 ,\\
\end{split}
\end{align}
and using the fact that $\mSigma$ is a symmetric matrix:
\begin{align}\label{eq:q_posterior_cov_derivation}
\begin{split}
    \tilde{\mSigma} &= 
    (1 - \bar\alpha_{\diffusionInd - 1}) \mSigma
    - 
    \frac{1 - \bar\alpha_{\diffusionInd - 1}}{1 - \bar\alpha_\diffusionInd} \sqrt{\alpha_\diffusionInd}
    \sqrt{1 - \beta_\diffusionInd}
    (1 - \bar\alpha_{\diffusionInd - 1}) \mSigma^T \\
    &= \left(
        1 - \bar\alpha_{\diffusionInd - 1}
        -
        \frac{1 - \bar\alpha_{\diffusionInd - 1}}{1 - \bar\alpha_\diffusionInd} \alpha_\diffusionInd
        (1 - \bar\alpha_{\diffusionInd - 1})
    \right) \mSigma \\
    &= \frac{1 - \bar\alpha_{\diffusionInd - 1}}{1 - \bar\alpha_\diffusionInd} \beta_\diffusionInd \mSigma .
\end{split}
\end{align}
Therefore, the only difference to the derivation in \citet{ho2020denoising} is the $\mSigma (\vt)$ instead of the identity matrix $\mI$ in the covariance.

\subsection{Discrete diffusion loss}\label{app:ddpm_loss}

We use the evidence lower bound from \Eqref{eq:elbo_ddpm}.
The distribution $q(\ts_{\diffusionInd - 1} | \ts_\diffusionInd, \ts_0)$ is defined as $\mathcal{N}(\tilde{\vmu}, C_1 \mSigma)$, where $C_1$ is some constant (Equations~\ref{eq:q_posterior_mean_derivation} and \ref{eq:q_posterior_cov_derivation}).
Similar to \citet{ho2020denoising}, we start with the parameterization for the reverse process $p(\ts_{\diffusionInd - 1} | \ts_\diffusionInd) = \mathcal{N}(\vmu_\theta(\ts_\diffusionInd, \vt, \diffusionInd), \beta_\diffusionInd \mSigma)$, where:
\begin{align*}
    \vmu_\theta(\ts_\diffusionInd, \vt, \diffusionInd) = \frac{1}{\sqrt{\alpha_\diffusionInd}} \left( \ts_\diffusionInd - \frac{\beta_\diffusionInd}{\sqrt{1 - \bar\alpha_\diffusionInd}} \noise_\theta(\ts_\diffusionInd, \vt, \diffusionInd) \right).
\end{align*}
Then the KL-divergence is between two normal distributions so we can write the following, where $C_2$ is a term that does not depend on the parameters $\theta$:
\begin{align*}
    \KL [ q(\ts_{\diffusionInd - 1} | \ts_\diffusionInd, \ts_0) \Ver p(\ts_{\diffusionInd - 1} | \ts_\diffusionInd) ] &= \KL [ \mathcal{N}(\tilde{\vmu}, C_1 \mSigma) \Ver \mathcal{N}(\vmu_\theta(\ts_\diffusionInd, \vt, \diffusionInd), \beta_\diffusionInd \mSigma)] \\
    &= \frac{1}{2} (\tilde\vmu - \vmu_\theta)^T \mSigma^{-1} (\tilde\vmu - \vmu_\theta) + C_2 .
\end{align*}
\citet{ho2020denoising} show that their loss can be simplified to \Eqref{eq:loss_ddpm} given their particular parameterization. Recall that we obtain noise by computing $\mL \tilde\noise$, where $\tilde\noise$ is unit normal and $\mL$ is the lower triangular matrix from the Cholesky decomposition of the covariance $\mSigma = \mL \mL^T$.

Therefore, we can factorize $\mL$ from the bracket containing the difference of two means to get:
\begin{align*}
    \KL [ q(\ts_{\diffusionInd - 1} | \ts_\diffusionInd, \ts_0) \Ver p(\ts_{\diffusionInd - 1} | \ts_\diffusionInd) ] = (\mL \va)^T \mSigma^{-1} (\mL \va) = \va^T \mL^T \mSigma^{-1} \mL \va,
\end{align*}
where we write $\va$ as a shorthand for the term depending on $\ts_0$ and unit normal noise $\tilde\noise$. The term $\mL^T \mSigma^{-1} \mL$ evaluates to identity and we are again left with the same loss as in \citet{ho2020denoising}.
That is, we can use the same trick to simplify the loss to be the mean squared error between the true noise and the predicted noise, which leads to faster evaluation and better results.
Note that in the above notation, we have a set of observations $\ts$ for times $\vt$ that we feed into the model $\noise_\theta$ to predict a set of noise values $\noise(t)$, $t \in \vt$, whereas, previous works predicted the noise for each data point independently.

\subsection{Continuous diffusion transition probability}\label{app:continuous_diffusion_transition}

Given an SDE in \Eqref{eq:ts_continuous_diffusion_sde} we want to compute the change in the variance $\tilde{\mSigma}_\diffusionContInd$, where $\diffusionContInd$ denotes the diffusion time. The derivation is similar to that in \citet{song2021sde}. We start with the Equation~5.51 from \citet{sarkka2019applied}:
\begin{align*}
     \frac{\diff \tilde{\mSigma}_\diffusionContInd}{\diff \diffusionContInd} = \E[
        f(\ts_\diffusionContInd, \diffusionContInd) (\ts_\diffusionContInd - \vmu)^T
     ] + \E[
        (\ts_\diffusionContInd - \vmu) f(\ts_\diffusionContInd, \diffusionContInd)^T
     ] + \E[
        \mL(\ts_\diffusionContInd, \diffusionContInd) \mQ \mL(\ts_\diffusionContInd, \diffusionContInd)^T
     ],
\end{align*}
where $f$ is the drift, $\mL$ is the SDE diffusion term and $\mQ$ is the diffusion matrix. From here, the only difference to \citet{song2021sde} is in the last term; they obtain $\beta(\diffusionContInd) \mI$ while we have a full covariance matrix from the stochastic process: $\beta(\diffusionContInd) \mSigma$. Therefore, we only need to slightly modify the result:
\begin{align*}
     \frac{\diff \mSigma_\diffusionContInd}{\diff \diffusionContInd} = \beta(\diffusionContInd) (\mSigma - \tilde{\mSigma}_\diffusionContInd) ,
\end{align*}
which will give us the covariance of the transition probability as in \Eqref{eq:ts_continuous_diffusion_transition}. The derivation for the mean is unchanged as our drift term is the same as in \citet{song2021sde}.

\subsection{Sampling from an Ornstein-Uhlenbeck process}\label{app:ou_sampling}

In the following, we discuss three different approaches to sampling noise $\noise(\cdot)$ from an OU process defined by $\gamma$ at time points $t_0, \dots, t_{\timeSeriesSize-1}$.
\begin{enumerate}
    \item \textbf{Modified Wiener.} As we already mentioned in \Secref{sec:ts_diffusion_noise}, we can use a time-changed and scaled Wiener process: $\smash{ e^{-\gamma t} W_{e^{2 \gamma  t}} }$. Sampling from a Wiener process is straightforward: given a set of time increments $\Delta t_0, \dots, \Delta t_{\timeSeriesSize-1}$, we sample $\timeSeriesSize$ points independently from $\mathcal{N}(0, \Delta t_i)$ and cumulatively sum all the samples. The time changed process first needs to reparameterize the time values. The issue arises when applying the exponential for large $t$ which leads to numerical instability. This can be mitigated by re-scaling $t$.
    \item \textbf{Discretized SDE.} A numerically stable approach involves \emph{solving} the OU SDE in fixed steps. The point at $t=0$, $\noise(0)$ is sampled from unit Gaussian. After that, each point is obtained based on the previous, i.e., $i$-th point $\noise(t_i)$ is calculated as $\noise(t_i) = c \noise(t_{i-1}) + \sqrt{1 - c^2} z$, where $c = \exp(-\gamma (t_{i} - t_{i-1}))$ and $z \sim \mathcal{N}(0, 1)$. This is an iterative procedure but is quite fast and stable.
    \item \textbf{Multivariate normal.} Finally, we can treat the process as a multivariate normal distribution with mean zero and covariance $\smash{\mSigma_{ij}(t_i, t_j) = \exp(-\gamma |t_i - t_j|)}$. Given a set of time points $\vt$ it is easy to obtain the covariance matrix $\mSigma$ and its factorization $\mL^T \mL$. To sample, we first draw $\tilde{\noise} \sim \mathcal{N}(\vzero, \mI)$ and then $\noise = \mL \tilde{\noise}$. Since our model performs best if it predicts $\tilde{\noise}$, we opted for this particular sampling approach. If $\vt$ is not changing, $\mL$ can be computed once and the performance impact will be minimal. Also when sampling new realizations, $\mL$ has to be computed only once, before the sampling loop (see Algorithm~\ref{alg:sampling_ddpm}).
\end{enumerate}

\subsection{Algorithms}\label{app:algorithms}

In Algorithms~\ref{alg:training_ddpm} and \ref{alg:sampling_ddpm} we provide the pseudocode for training the model and sampling new data, for DSPD-GP model. Analogously for OU, we would replace the noise source using the third algorithm from \Appref{app:ou_sampling}. For the score-based model we compute the mean squared error between the predicted and true conditional score function and the sampling uses either ODE or SDE solver, just like in \citet{song2021sde}.

\begin{minipage}{0.45\textwidth}
    \begin{algorithm}[H]
        \centering
        \caption{Training (DSPD-GP diffusion)}\label{alg:training_ddpm}
        \footnotesize
        \begin{algorithmic}[1]
            \While{not converged}
                \State $\ts_0, \vt \sim p_\text{data}(\ts, \vt)$
                \State $\mSigma = k(\vt, \vt^T)$
                \State $\mL = \mathrm{Cholesky}(\mSigma)$
                \State $\tilde{\noise} \sim \mathcal{N}(\vzero, \mI)$
                \State $\diffusionInd \sim \mathcal{U}(\{ 1, \dots, \diffusionLast \})$
                \State $\ts_\diffusionInd = \sqrt{\alpha_\diffusionInd} \ts_0 + \sqrt{1 - \alpha_\diffusionInd} \mL \tilde{\noise}$
                \State Take gradient step on
                \State \indent $\nabla_\theta || \tilde\noise - \noise_\theta(\ts_\diffusionInd, \vt, \diffusionInd) ||_2^2$
            \EndWhile
        \end{algorithmic}
    \end{algorithm}
\end{minipage}
\hfill
\begin{minipage}{0.5\textwidth}
    \begin{algorithm}[H]
        \centering
        \caption{Sampling (DSPD-GP diffusion)}\label{alg:sampling_ddpm}
        \footnotesize
        \begin{algorithmic}[1]
        \State \textbf{input:} $\vt = ( t_0, \dots, t_{\timeSeriesSize-1} )$
        \State $\mSigma = k(\vt, \vt^T)$; $\mL = \mathrm{Cholesky}(\mSigma)$
        \State $\ts_\diffusionLast \sim \mathcal{N}(\vzero, \mSigma)$
        \For{$\diffusionInd = \diffusionLast, \dots, 1$}
            \State $\vz \sim \mathcal{N}(\vzero, \mSigma)$
            \State $\smash{
            \ts_{\diffusionInd - 1} = \frac{1}{\sqrt{\alpha_\diffusionInd}} \left( \ts_\diffusionInd - \frac{1 - \alpha_\diffusionInd}{\sqrt{1 - \bar\alpha_\diffusionInd}}
            \mL \noise_\theta(\ts_\diffusionInd, \vt, \diffusionInd)
            \right) + \beta_\diffusionInd \vz
            }$
        \EndFor
        \State \textbf{return} $\ts_0$
        \end{algorithmic}
    \end{algorithm}
\end{minipage}

\section{Experimental details}\label{app:results}

\subsection{Probabilistic modeling}\label{app:probabilistic_modeling}

\subsubsection{Datasets} 
The properties of the open datasets used in the forecasting experiment are detailed in Table \ref{table:dataset}. Additionally, we generate 6 synthetic datasets, each with 10000 samples, that involve stochastic processes, dynamical and chaotic systems.
\begin{enumerate}
    \item CIR (Cox-Ingersoll-Ross SDE) is the stochastic differential equations defined by:
    \begin{align*}
        \diff x = a (b - x) \diff t + \sigma \sqrt{x} \diff W_t,
    \end{align*}
    where we set $a=1$, $b=1.2$, $\sigma=0.2$ and sample $x_0 \sim \mathcal{N}(0, 1)$ but only take the positive values, otherwise the $\sqrt{x}$ term is undefined. We solve for $t \in \{1, \dots, 64\}$.
    \item Lorenz is a chaotic system in three dimensions. It is governed by the following equations:
    \begin{align*}
        \dot{x} &= \sigma (y - x), \\
        \dot{y} &= \rho x - y - x  z, \\
        \dot{z} &= x y - \beta z, \\
    \end{align*}
    where $\rho=28$, $\sigma=10$, $\beta=2.667$, and $t$ is sampled 100 times, uniformly on $[0, 2]$, and $x, y, z \sim \mathcal{N}(\vzero, 100 \mI)$.
    \item Ornstein-Uhlenbeck is defined as:
    \begin{align*}
        \diff x = (\mu t - \theta x) \diff t + \sigma \diff W_t,
    \end{align*}
    with $\mu=0.02$, $\theta=0.1$ and $\sigma=0.4$. We sample time the same way as for CIR.
    \item Predator-prey is a 2D dynamical system defined with an ODE:
    \begin{align*}
        \dot{x} &= 2/3 x - 2/3 x y, \\
        \dot{y} &= x y - y.
    \end{align*}
    \item Sine dataset is generated as a mixture of 5 random sine waves $a \sin(b x + c)$, where $a \sim \mathcal{N}(3, 1)$, $b \sim \mathcal{N}(0, 0.25)$, and $c \sim \mathcal{N}(0, 1)$.
    \item Sink is again a dynamical system, governed by:
    \begin{align*}
        \frac{\diff \vx}{\diff t} = \begin{bmatrix}
            -4 & 10 \\
            -3 & 2
        \end{bmatrix} \vx ,
    \end{align*}
    with $\vx_0 \sim \mathcal{N}(\vzero, \mI)$.
\end{enumerate}

\begin{table}[t]
\vskip 0.15in
\caption{Multivariate dimension, domain, frequency, total training time steps, and prediction length properties of the  training datasets used in the forecasting experiments.}
\label{table:dataset}
\begin{center}
\begin{small}
\begin{tabular}{lccccc}
Dataset &   Dim.\ $d$ & Dom. & Freq. & Time steps  & Pred.\ steps 
\\ 
\midrule
Exchange  & $8$ & $\mathbb{R}^{+}$ & day & $6,071$ & $30$ 
\\
Solar    & $137$ & $\mathbb{R}^{+}$ & hour & $7,009$ & $24$ 
\\
Electricity & $370$ &  $\mathbb{R}^{+}$ & hour & $5,833$ & $24$ 
 \\
\end{tabular}

\end{small}
\end{center}
\vskip -0.1in
\end{table}

\subsubsection{CTFP} 
We implement continuous-time flow process \citep{deng2020modeling} which is a normalizing flow model for stochastic processes. That is, there is a predefined base distribution $p(\vz)$ and a series of invertible transformations $f$ such that we can generate samples $\vx = f(\vz)$, and evaluate the density in closed-form by computing $\vz = f^{-1}(\vx)$ and using the change of variables formula. For more details on normalizing flows, see \citet{kobyzev2020normalizing}. The novel idea in CTFP is to change the base density to a stochastic process, i.e., a Wiener process, to obtain the distribution over the functions, similar to our work. In our case, we do not use invertible functions but learn to inverse the noising process, and additionally, we add noise at multiple levels instead only in the beginning. In the experiments, we define a CTFP model as a 12-layer real NVP architecture \citep{realnvp} with 2 hidden layers in each layer's MLP.

\subsubsection{Latent ODE} 
Latent ODE is a variational autoencoder architecture, with an encoder that represents the complete time series as a single vector following $q(\vz)$, and a decoder that produces the samples at observation times $t_i$, $\vz(t_i) = f(\vz), \vz \sim q(\vz)$. The final step is projection to a data space $\vq(t_i) \mapsto \vx(t_i)$. The key idea is to use the neural ordinary differential equation \citep{chen2018neural} to define the evolution of the latent variable $\vz(\cdot)$, thus, have a probabilistic model of the function. This is different from our approach as it models the function in a latent space, with a single source of randomness at the beginning of the time series. That is, the random value is sampled at $t=0$ and the time series is determined from there onward, whereas our method samples random values on the whole interval $[0, T]$ and does so multiple times (for $\diffusionLast$ diffusion steps) until we get the new realization. In the experiments, we use a two layer neural network for the neural ODE, and another two layer network for projection to the data space.

\subsubsection{Our models} 
We use two models, one is a simple feedforward network, and the second is an RNN-based model. We also use a simple transformer-based model \cite{vaswani2017attention} that achieves similar results to an RNN. The model takes in the time series $\ts$, times of the observations $\vt$ and the diffusion step $\diffusionInd$ or diffusion time $\diffusionContInd$. The output is the same size as $\ts$. The feedforward model embeds the time and the diffusion step with a positional encoding \citep{vaswani2017attention} and passes it together with $\ts$ through the multilayer neural network. Here, there is no interaction between the points along the time dimension. The model, however, has the capacity to learn transformation based on time of observation. The second model is RNN based, that is, we pass the same concatenated input as before to a 2-layer bidirectional GRU \citep{gru} and use a single linear layer to project to the output dimension. \Tabref{tab:further_discriminator_results} shows that it is important to have interactions in the time dimension, regardless of the noise source, because otherwise we only learn the marginal distribution and the quality of the samples suffers.

\begin{table*}[t]
    \centering
    \caption{Negative log-likelihood on synthetic data (lower is better) shows OU/GP is mostly better than independent noise.}
    \label{tab:synthetic_results}
    \vspace{0.1cm}
    \begin{adjustbox}{width=\textwidth}
        \begin{tabular}{lcccccc}
 &                 CIR &              Lorenz &                 OU &       Predator-prey &                Sine &                 Sink \\
\midrule
\citet{song2021sde} &  -0.4769$\pm$0.0249 &   1.5162$\pm$0.3861 &  0.5105$\pm$0.0088 &  -3.4643$\pm$0.1039 &  -1.3338$\pm$0.0863 &   -5.6637$\pm$0.1839 \\
CSPD-GP &  -0.4766$\pm$0.0224 &  -3.4893$\pm$8.2181 &  0.5202$\pm$0.0255 &  -9.4478$\pm$0.2466 &  -3.4878$\pm$1.3467 &  -11.4179$\pm$0.3627 \\
CSPD-OU &  -0.4688$\pm$0.0178 &   -6.6707$\pm$0.175 &  0.5239$\pm$0.0639 &  -7.0098$\pm$1.4929 &  -3.5324$\pm$0.6466 &   -9.5349$\pm$1.3183 \\
\end{tabular}

    \end{adjustbox}
\end{table*}
\begin{table}[t]
    \caption{Accuracy of the discriminator trained on samples from a diffusion model. Values around 0.5 indicate the discriminative model cannot distinguish the model samples and real data. Values closer to 1 indicate the generative model is not capturing the data distribution.}
    \label{tab:further_discriminator_results}
    \centering
    \resizebox{\textwidth}{!}{
    \begin{tabular}{llcccccc}
    && CIR & Lorenz & OU & Predator-prey & Sine & Sink \\
    \multicolumn{8}{l}{RNN-based model} \\
    \midrule
    \multirow{3}{*}{\rotatebox[origin=c]{90}{DSPD}} & 
    Gauss &  0.5245$\pm$0.0252 &   0.512$\pm$0.0212 &    0.568$\pm$0.051 &  0.5275$\pm$0.0383 &  0.5565$\pm$0.0353 &   0.526$\pm$0.0085 \\
    & GP &  0.5115$\pm$0.0282 &  0.5135$\pm$0.0288 &  0.5055$\pm$0.0458 &  0.5855$\pm$0.0219 &   0.5255$\pm$0.009 &   0.513$\pm$0.0103 \\
    & OU &   0.514$\pm$0.0737 &  0.6095$\pm$0.0964 &  0.5605$\pm$0.0581 &   0.5865$\pm$0.053 &     0.507$\pm$0.11 &  0.6255$\pm$0.1672 \\
    \multirow{3}{*}{\rotatebox[origin=c]{90}{CSPD}} & Gauss &   0.644$\pm$0.0373 &  0.5015$\pm$0.0243 &  0.6105$\pm$0.0153 &   0.548$\pm$0.0751 &   0.611$\pm$0.0516 &  0.5495$\pm$0.0313 \\
    &  GP &  0.5795$\pm$0.0541 &   0.674$\pm$0.0739 &  0.5025$\pm$0.0622 &   0.607$\pm$0.0538 &  0.5575$\pm$0.0376 &  0.5345$\pm$0.0201 \\
    & OU &   0.4535$\pm$0.165 &   0.715$\pm$0.0884 &   0.5255$\pm$0.011 &  0.5835$\pm$0.0723 &    0.556$\pm$0.118 &  0.5795$\pm$0.0173 \\
    \\
    \multicolumn{8}{l}{Feedforward model} \\
    \midrule
    \multirow{3}{*}{\rotatebox[origin=c]{90}{DSPD}} 
    & Gauss &  0.624$\pm$0.0438 &  0.713$\pm$0.1798 &  0.5275$\pm$0.0371 &        1.0$\pm$0.0 &  0.7875$\pm$0.0585 &  0.9695$\pm$0.0302 \\
    & GP &  0.558$\pm$0.0611 &   0.894$\pm$0.212 &  0.5535$\pm$0.1152 &  0.7565$\pm$0.1362 &   0.735$\pm$0.2146 &   0.784$\pm$0.2281 \\
    & OU&       1.0$\pm$0.0 &       1.0$\pm$0.0 &        1.0$\pm$0.0 &        1.0$\pm$0.0 &        1.0$\pm$0.0 &        1.0$\pm$0.0 \\
    \multirow{3}{*}{\rotatebox[origin=c]{90}{CSPD}} & Gauss &  0.537$\pm$0.0458 &  0.959$\pm$0.0808 &  0.5155$\pm$0.0165 &   0.9995$\pm$0.001 &  0.6335$\pm$0.0765 &  0.9095$\pm$0.1306 \\
    & GP &  0.645$\pm$0.1034 &       1.0$\pm$0.0 &   0.507$\pm$0.0264 &    0.894$\pm$0.212 &    0.894$\pm$0.212 &     0.88$\pm$0.088 \\
    & OU       &   0.984$\pm$0.032 &       1.0$\pm$0.0 &   0.9905$\pm$0.019 &        1.0$\pm$0.0 &        1.0$\pm$0.0 &        1.0$\pm$0.0 \\
    \end{tabular}
    }

\end{table}

\subsection{Multivariate probabilistic forecasting}\label{app:forecasting}

\begin{table}[h!]
    \centering
    \caption{CRPS-sum results on forecasting task. Values for non-diffusion baselines are taken from \citet{salinas2019high}.}
    \label{tab:forecasting_crps}
    \begin{tabular}{lccc}
                            & Electricity      & Exchange           & Solar \\
        \midrule
        LSTM                & 0.025$\pm$0.001  & 0.008$\pm$0.001 & 0.391$\pm$0.017\\
        LSTM-Copula         & 0.064$\pm$0.008  & 0.007$\pm$0.000 & 0.319$\pm$0.011\\
        GP                  & 0.947$\pm$0.016  & 0.011$\pm$0.001 & 0.828$\pm$0.010\\
        GP-Copula           & 0.024$\pm$0.002  & 0.007$\pm$0.000 & 0.337$\pm$0.024\\
        \midrule
        TimeGrad            &  0.036$\pm$0.002 & 0.009$\pm$0.001 &  0.389$\pm$0.041 \\
        Our                 &  0.027$\pm$0.001 & 0.007$\pm$0.001 &  0.371$\pm$0.034 \\
    \end{tabular}
\end{table}

\subsection{Neural process}\label{app:neural_process}

\subsubsection{Dataset} 
We sample points from a Gaussian process to obtain a single time series. In the end, we have 8000 time series and 2000 test time series. We sample the number of time points from a Poisson distribution with $\lambda=10$ but restrict the values to always be above 5 and below 50. The time points are sampled uniformly on $[0, 1]$. The observations are sampled from a multivariate normal distribution with mean zero and covariance obtained from an RBF kernel. The $\sigma$ value in the kernel is uniformly sampled in $[0.01, 0.05]$ for each time series independently. Half of the sampled points are treated as unobserved while the rest are used as a context in the model.

\subsubsection{Model} 
The denoising model takes in $\ts^A$ (observed points) as a conditioning variable and $\ts^B_\diffusionInd$ (target points) as the noisy input. We first run a learnable RBF kernel $k(\vt^A, \vt^B)$ to obtain a similarity matrix $\mK$ between the observed and unobserved time points. We project $\ts^A$ with a neural network by transforming each point independently to obtain $\mZ$, and then obtain the latent variable of the same time dimension size as $\ts^B$ by multiplying $\mK$ and $\mZ$. We then use $\mZ$ as a conditioning vector and add it to projected $\ts^B$, transform with a multilayer network, and obtain the output.

\subsubsection{Additional results}

\begin{figure}
    \centering
    \resizebox{0.6 \linewidth}{!}{
        \input{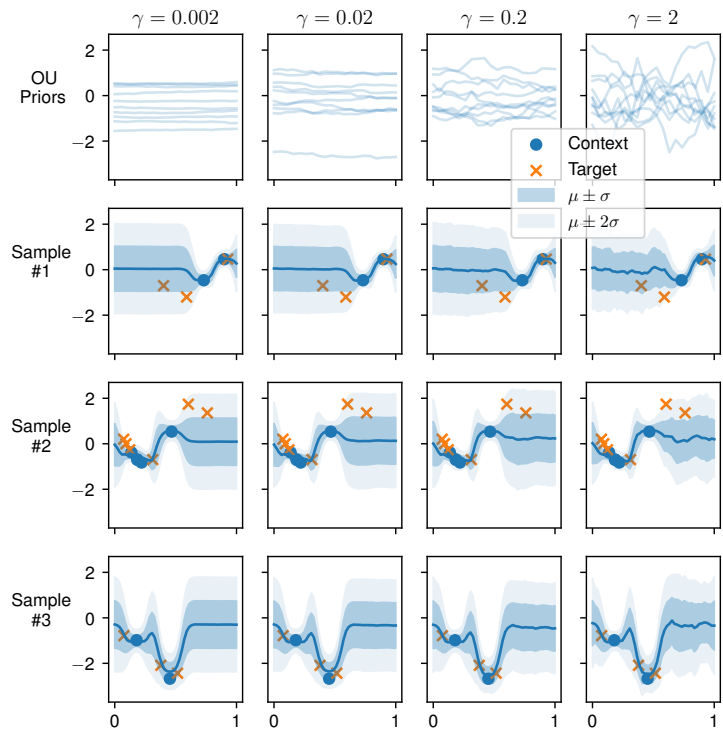}
    }
    \caption{Same setting as \Figref{fig:gp_np} but for the Ornstein-Uhlenbeck process. Here, increasing the kernel parameter $\gamma$ now decreases the smoothness. All of the models perfectly capture the marginal distribution.}
    \label{fig:ou_np}
\end{figure}

We test the hypothesis that using a stochastic process with similar properties to the data will lead to better performance. The difference to the neural process setup in \Secref{sec:experiments} is that we fix the synthetic GP to always have $\sigma=0.05$. As can be seen from Figures~\ref{fig:gp_np} and \ref{fig:ou_np}, the marginal distribution will be equal regardless of which process and which kernel parameter we use. On the other hand, when we look at path probability $p(\ts)$, we notice better results when the noise process matches data properties (as was also shown in \Tabref{tab:synthetic_results} and \ref{tab:further_discriminator_results}). That means, while our model can reverse the process well, the qualitative properties of the sampled curves will be different. In particular, the curves will be \textit{rougher} with increasing $\gamma$ in OU and \textit{smoother} with increasing $\sigma$ in GP.

\subsection{CSDI imputation}\label{app:imputation}

The imputation experiment presented in Sections~\ref{sec:imputation}~and~\ref{sec:experiments} uses the original CSDI model \citep{tashiro2021csdi} and only changes the noise to include the stochastic process source. In this case, the time points at which we evaluate the stochastic process are regular which does not reflect the true nature of the Physionet dataset. Here, we change the setup such that the measurements keep the actual time that has passed instead of rounding to the nearest hour. This is still in favour of the original paper as it only takes one measurement per hour and discards others if they are present. The model from \citet{tashiro2021csdi} remains the same and we replace the independent normal noise with the GP noise with 
$\sigma \in \{ 0.005, 0.01, 0.02 \}$.

We run each experimental setup 10 times with different data maskings (see \citet{tashiro2021csdi} for more details) and report the results in \Tabref{tab:csdi-additional}. We perform the Wilcoxon one-sided signed-rank test \citep{conover1999practical} and reject the null hypothesis that the expected RMSE values are the same when $p < 0.05$. As we can see, higher values of $\sigma$ produce better results which makes sense since $\sigma = 0.005$ is, informally, closer to independent Gaussian sampling than $\sigma =0.02$, which has stronger temporal dependency between the samples. We suspect 10\%-missing case does not produce significant results due to noise. Using higher $\sigma$ does not further improve the results.

\begin{table}[h]
    \caption{Imputation results averaged over 10 runs and p-value of Wilcoxon one-sided test.}
    \label{tab:csdi-additional}
    \centering
    \begin{tabular}{llcccccc}
    \multicolumn{2}{l}{Missingness:} & \multicolumn{2}{c}{10\%} & \multicolumn{2}{c}{50\%} & \multicolumn{2}{c}{90\%} \\
    \multicolumn{2}{l}{Metrics:} & RMSE & p-value & RMSE & p-value & RMSE & p-value \\
    \midrule
    \multicolumn{2}{l}{CSDI (baseline)} &  0.603$\pm$0.274 & -- & 0.658$\pm$0.060 & -- & 0.839$\pm$0.043 & -- \\
    \midrule
    & 0.005 & 0.541$\pm$0.085 & 0.125 & 0.647$\pm$0.049 & 0.116 & 0.824$\pm$0.032 & 0.188 \\
    $\sigma=$ & 0.01  & 0.575$\pm$0.195 & 0.125 & 0.640$\pm$0.050 & \textbf{0.001} & 0.823$\pm$0.028 & \textbf{0.032} \\
    & 0.02  & 0.515$\pm$0.039 & 0.326 & 0.636$\pm$0.050 & \textbf{0.001} & 0.811$\pm$0.032 & \textbf{0.001} \\
\end{tabular}
\end{table}

\end{document}